\documentclass[10pt, a4paper, conference, compsocconf]{IEEEtran}

\usepackage{times}
\usepackage{balance}
\usepackage{setspace}
\usepackage{algorithm}
\usepackage{algorithmic}
\usepackage{boxedminipage} 
\usepackage{graphicx}
\usepackage{verbatim} 
\usepackage{amsmath}
\usepackage{times} 
\usepackage{gensymb}
\usepackage{dsfont}
\usepackage{multirow}
\usepackage{amsfonts}
\usepackage{float}
\usepackage{epstopdf}
\usepackage[T1]{fontenc}
\usepackage{url}
\usepackage{amssymb}
\usepackage[tight,footnotesize]{subfigure}
\usepackage[caption=false,font=footnotesize]{subfig}
\usepackage{epsfig}
\usepackage{bm}
\begin{document}
%
\title{META-DES.H: a dynamic ensemble selection technique using meta-learning and a dynamic weighting approach}

\author{\IEEEauthorblockN{Rafael M. O. Cruz and Robert Sabourin}
\IEEEauthorblockA{\'{E}cole de technologie sup\'{e}rieure - Universit\'{e} du Qu\'{e}bec\\
Email: cruz@livia.etsmtl.ca, robert.sabourin@etsmtl.ca}
\and
\IEEEauthorblockN{George D. C. Cavalcanti}
\IEEEauthorblockA{Centro de Inform\'{a}tica - Universidade Federal de Pernambuco\\
Email: gdcc@cin.ufpe.br}
}

\maketitle

\begin{abstract}

In Dynamic Ensemble Selection (DES) techniques, only the most competent classifiers are selected to classify a given query sample. Hence, the key issue in DES is how to estimate the competence of each classifier in a pool to select the most competent ones. In order to deal with this issue, we proposed a novel dynamic ensemble selection framework using meta-learning, called META-DES. The framework is divided into three steps. In the first step, the pool of classifiers is generated from the training data. In the second phase the meta-features are computed using the training data and used to train a meta-classifier that is able to predict whether or not a base classifier from the pool is competent enough to classify an input instance. In this paper, we propose improvements to the training and generalization phase of the META-DES framework. In the training phase, we evaluate four different algorithms for the training of the meta-classifier. For the generalization phase, three combination approaches are evaluated: Dynamic selection, where only the classifiers that attain a certain competence level are selected; Dynamic weighting, where the meta-classifier estimates the competence of each classifier in the pool, and the outputs of all classifiers in the pool are weighted based on their level of competence; and a hybrid approach, in which first an ensemble with the most competent classifiers is selected, after which the weights of the selected classifiers are estimated in order to be used in a weighted majority voting scheme. Experiments are carried out on 30 classification datasets. Experimental results demonstrate that the changes proposed in this paper significantly improve the recognition accuracy of the system in several datasets.

\end{abstract}


\section{Introduction}

Multiple Classifier Systems (MCS) aim to combine classifiers in order to increase the recognition accuracy in pattern recognition systems~\cite{kittler,kuncheva}. MCS are composed of three phases~\cite{Alceu2014}: (1) Generation, (2) Selection, and (3) Integration. In the first phase, a pool of classifiers is generated. In the second phase, a single classifier or a subset having the best classifiers of the pool is(are) selected. We refer to the subset of classifiers as the Ensemble of Classifiers (EoC). In the last phase, integration, the predictions of the selected classifiers are combined to obtain the final decision~\cite{kittler}.

Recent works in MCS have shown that dynamic ensemble selection (DES) techniques achieve higher classification accuracy when compared to static ones~\cite{Alceu2014,CruzPR,knora}. This is specially true for ill-defined problems, i.e., for problems where the size of the training data is small and there are not enough data available to train the classifiers~\cite{paulo2,logid}. The key issue in DES is to define a criterion to measure the level of competence of a base classifier. Most DES techniques~\cite{knora,lca,mcb,ijcnn2011} use estimates of the classifiers' local accuracy in small regions of the feature space surrounding the query instance, called the region of competence, as a search criterion to estimate the competence level of the base classifier. However, in our previous work~\cite{ijcnn2011}, we demonstrated that the use of local accuracy estimates alone is insufficient to provide higher classification performance. 

To tackle this issue, in~\cite{CruzPR} we proposed a novel DES framework, called META-DES, in which multiple criteria regarding the behavior of a base classifier are used to compute its level of competence. The framework is based on two environments: the classification environment, in which the input features are mapped into a set of class labels, and the meta-classification environment, where different properties from the classification environment, such as the classifier accuracy in a local region of the feature space, are extracted from the training data and encoded as meta-features. With the arrival of new test data, the meta-features are extracted using the test data as reference, and used as input to the meta-classifier. The meta-classifier decides whether the base classifier is competent enough to classify the test sample. The framework is divided into three steps: (1) Overproduction, where the pool of classifiers is generated; (2) Meta-training, where the meta-features are extracted, using the training data, and used as inputs to train a meta-classifier that works as a classifier selector, and (3) the Generalization phase, in which the meta-features are extracted from each query sample and used as input to the meta-classifier to perform the ensemble selection.

In this paper, we propose two improvements to the META-DES framework. First, we modify the training routine of the meta-classifier. The modification made is motivated by the fact that there is a strong correlation between the performance of the meta-classifier for the selection of ``competent'' classifiers, i.e., classifiers that predict the correct label for a given query sample and the classification accuracy of the DES system~\cite{icpr2014}. Hence, we believe that the proposed META-DES framework can obtain higher classification performance by focusing only on improving the performance of the system at the meta-classification level. This is an interesting feature of the proposed system especially when dealing with ill-defined problems due to critical dataset sizes~\cite{CruzPR}. Four different classifier models are considered for the meta-classifier: MLP Neural Network, Support Vector Machines with Gaussian Kernel (SVM), s and Naive Bayes~\cite{delgado14a}.

Secondly, we propose three combination schemes for the generalization phase of the framework: Dynamic selection, Dynamic weighting and Hybrid. In the dynamic selection approach, only the classifiers that attain a certain level of competence are used to classify a given query sample. In the dynamic weighting approach, the meta-classifier is used to estimate the weights of all base classifiers in the pool. Then, their decisions are aggregated using a weighted majority voting scheme~\cite{kuncheva}. Thus, classifiers that attain a higher level of competence, for the classification of the given query sample, have a greater impact on the final decision. In the hybrid approach, only the classifiers that attain a certain level of competence are selected. Then, the meta-classifier is used to compute the weights of the selected base classifiers to be used in a weighted majority voting scheme. The hybrid approach is based on the observation that the selected base classifiers might be associated with different levels of competence. It is feasible that classifiers that attained a higher level of competence should have more influence for the classification of the given test sample. The proposed framework differs from mixture of expert techniques~\cite{NguyenAM06,mixture}, since our system is based on the mechanism used for the selection of dynamic ensembles~\cite{Alceu2014,CruzPR} rather than static ones~\cite{mixture}. In addition, mixture of experts techniques are dedicated to the use of neural networks as base classifier, while, in the proposed framework, any classification algorithm can be used.

We evaluate the generalization performance of the system over 30 classification problems derived from different data repositories. Furthermore, the recognition performance of the system is compared against eight state-of-the-art dynamic selection techniques according to a new survey on this topic~\cite{Alceu2014}. Experimental results demonstrate that the choice of the meta-classifier has a significant impact on the classification accuracy of the overall system. The modifications proposed in this work significantly improve the performance of the framework when compared to state-of-the-art dynamic selection techniques.

This paper is organized as follows: The META-DES framework is introduced in Section~\ref{sec:proposed}. Experimental results are given in Section~\ref{sec:experiments}. Finally the conclusion is presented in the last section.

\section{The META-DES Framework}
\label{sec:proposed}

The META-DES framework is based on the assumption that the dynamic ensemble selection problem can be considered as a meta-problem. This meta-problem uses different criteria regarding the behavior of a base classifier $c_{i}$, in order to decide whether it is competent enough to classify a given test sample $\mathbf{x}_{j}$. The meta-problem is defined as follows~\cite{CruzPR}:

 \begin{itemize}
 
 \item The \textbf{meta-classes} of this meta-problem are either ``competent'' (1) or ``incompetent'' (0) to classify $\mathbf{x}_{j}$.
 
 \item Each set of \textbf{meta-features} $f_{i}$ corresponds to a different criterion for measuring the level of competence of a base classifier.
 
 \item The meta-features are encoded into a \textbf{meta-features vector} $v_{i,j}$.
 
 \item A \textbf{meta-classifier} $\lambda$ is trained based on the meta-features $v_{i,j}$ to predict whether or not $c_{i}$ will achieve the correct prediction for $\mathbf{x}_{j}$, i.e., if it is competent enough to classify $\mathbf{x}_{j}$
 
 \end{itemize}
 
\begin{figure*}[!ht]
  
   \begin{center}  	 
       	  \epsfig{file=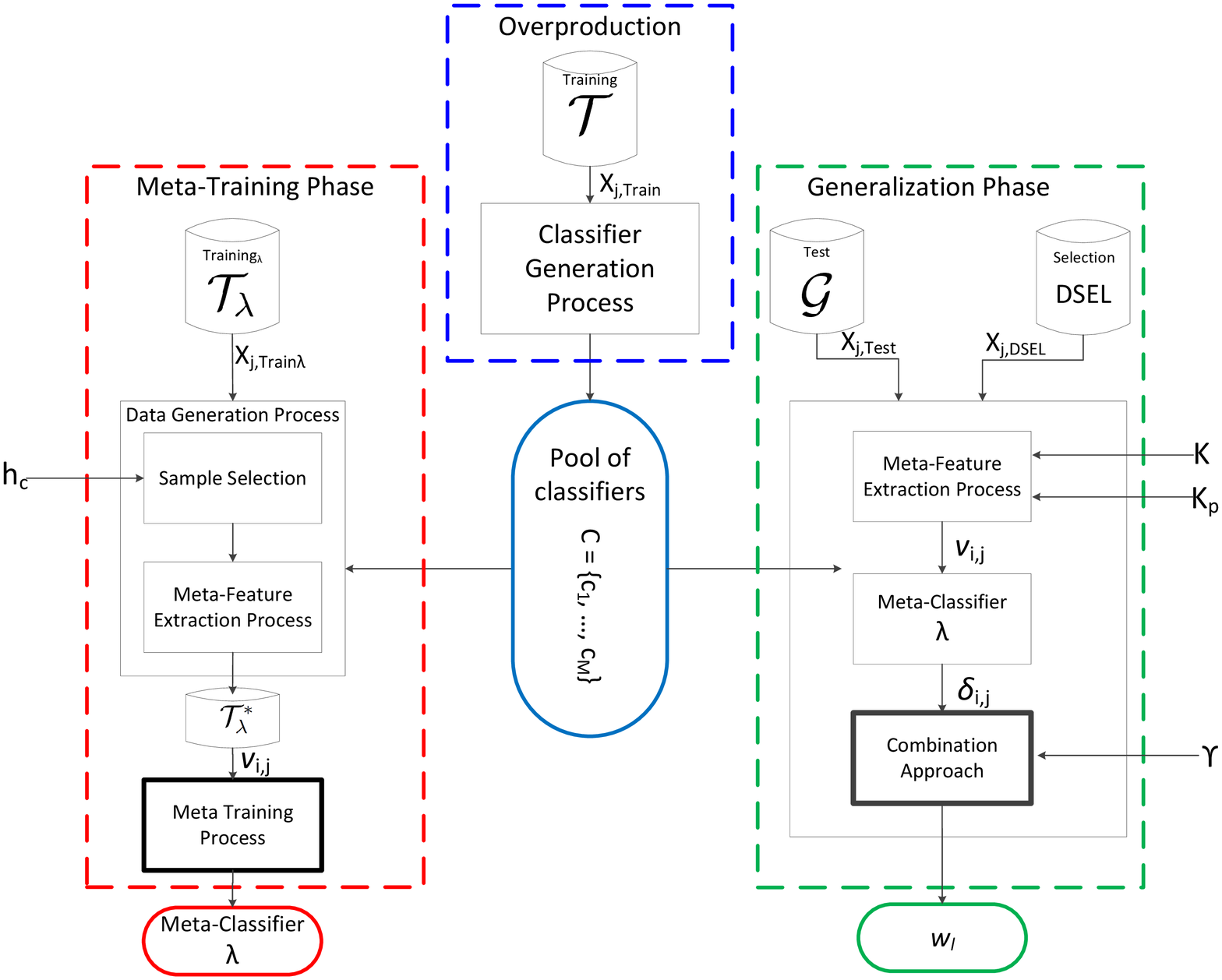, clip=,  width=0.900\textwidth}
   \end{center}
\caption{Overview of the proposed framework. It is divided into three steps 1) Overproduction, where the pool of classifiers $C = \{c_{1}, \ldots, c_{M}\}$ is generated, 2) The training of the selector $\lambda$ (meta-classifier), and 3) The generalization phase where the level of competence $\delta_{i,j}$ of each base classifier $c_{i}$ is calculated specifically for each new test sample $\mathbf{x}_{j,test}$. Then, the level of competence $\delta_{i,j}$ is used by the combination approach to predict the label $w_{l}$ of the test sample $\mathbf{x}_{j,test}$. Three combination approaches are considered: Dynamic selection (META-DES.S), Dynamic weighting (META-DES.W) and Hybrid (META-DES.H). $h_{C}$, $K$, $K_{p}$ and $\Upsilon$ are the hyper-parameters required by the proposed system. [Adapted from~\cite{CruzPR}].}
\label{fig:overview}
\end{figure*}

A general overview of the META-DES framework is depicted in Figure~\ref{fig:overview}. It is divided into three phases: Overproduction, Meta-training and Generalization.

\subsection{Overproduction} 

In this step, the pool of classifiers $C = \{c_{1}, \ldots, c_{M}\}$, where $M$ is the pool size, is generated using the training dataset $\mathcal{T}$. The Bagging technique~\cite{bagging} is used in this work in order to build a diverse pool of classifiers. 

\subsection{Meta-Training}

In this phase, the meta-features are computed and used to train the meta-classifier $\lambda$. As shown in Figure~\ref{fig:overview}, the meta-training stage consists of three steps: sample selection, meta-features extraction process and meta-training. A different dataset $\mathcal{T}_{\lambda}$ is used in this phase to prevent overfitting.

\subsubsection{Sample selection}
 
We decided to focus the training of $\lambda$ on cases in which the extent of consensus of the pool is low. This decision was based on the observations made in~\cite{docs,paulo2} the main issues in dynamic ensemble selection occur when classifying testing instances where the degree of consensus among the pool of classifiers is low, i.e., when the number of votes from the winning class is close to or even equal to the number of votes from the second class. We employ a sample selection mechanism based on a threshold $h_{C}$, called the consensus threshold. For each $\mathbf{x}_{j,train_{\lambda}} \in \mathcal{T}_{\lambda}$, the degree of consensus of the pool, denoted by $H \left ( \mathbf{x}_{j,train_{\lambda}}, C \right )$, is computed. If $H \left ( \mathbf{x}_{j,train_{\lambda}}, C \right )$ falls below the threshold $h_{C}$, $\mathbf{x}_{j,train_{\lambda}}$ is passed down to the meta-features extraction process. 

\subsubsection{Meta-feature extraction}
\label{sec:metafeatures}

The first step in extracting the meta-features involves computing the region of competence of $\mathbf{x}_{j,train_{\lambda}}$, denoted by $\theta_{j} = \left \{ \mathbf{x}_{1}, \ldots, \mathbf{x}_{K} \right \}$. The region of competence is defined in the $\mathcal{T_{\lambda}}$ set using the K-Nearest Neighbor algorithm. Then, $\mathbf{x}_{j,train_{\lambda}}$ is transformed into an output profile, $\tilde{\mathbf{x}}_{j,train_{\lambda}}$. The output profile of the instance $\mathbf{x}_{j,train_{\lambda}}$ is denoted by $\tilde{\mathbf{x}}_{j,train_{\lambda}} = \left\lbrace \tilde{\mathbf{x}}_{j,train_{\lambda},1}, \tilde{\mathbf{x}}_{j,train_{\lambda},2}, \ldots, \tilde{\mathbf{x}}_{j,train_{\lambda},M} \right\rbrace $, where each $\tilde{\mathbf{x}}_{j,train_{\lambda},i}$ is the decision yielded by the base classifier $c_{i}$ for the sample $\mathbf{x}_{j,train_{\lambda}}$~\cite{paulo2}.

The similarity between $\tilde{\mathbf{x}}_{j,train_{\lambda}}$ and the output profiles of the instances in $\mathcal{T}_{\lambda}$ is obtained through the Euclidean distance. The most similar output profiles are selected to form the set $\phi_{j} = \left \{ \tilde{\mathbf{x}}_{1}, \ldots, \tilde{\mathbf{x}}_{K_{p}} \right \}$, where each output profile $\tilde{\mathbf{x}}_{k}$ is associated with a label $w_{l,k}$. Next, for each base classifier $c_{i} \in C$, five sets of meta-features are calculated:

\begin{itemize}

\item  \emph{\boldsymbol{$f_{1}$} \textbf{- Neighbors' hard classification:}} First, a vector with $K$ elements is created. For each sample $\mathbf{x}_{k}$, belonging to the region of competence $\theta_{j}$, if $c_{i}$ correctly classifies $\mathbf{x}_{k}$, the $k$-th position of the vector is set to 1, otherwise it is 0. Thus, $K$ meta-features are computed. 

\item \emph{\boldsymbol{$f_{2}$} \textbf{- Posterior Probability:}} First, a vector with $K$ elements is created. Then, for each sample $\mathbf{x}_{k}$, belonging to the region of competence $\theta_{j}$, the posterior probability of $c_{i}$, $P(w_{l}\mid \mathbf{x}_{k})$ is computed and inserted into the $k$-th position of the vector. Consequently, $K$ meta-features are computed. 

\item \emph{\boldsymbol{$f_{3}$} \textbf{- Overall Local Accuracy:}} The accuracy of $c_{i}$ over the whole region of competence $\theta_{j}$ is computed and encoded as $f_{3}$. 

\item \emph{\boldsymbol{$f_{4}$} \textbf{- Outputs' profile classification:}} First, a vector with $K_{p}$ elements is generated. Then, for each member $\tilde{\mathbf{x}}_{k}$ belonging to the set of output profiles $\phi_{j}$, if the label produced by $c_{i}$ for $\mathbf{x}_{k}$ is equal to the label $w_{l,k}$ of $\tilde{\mathbf{x}}_{k}$, the $k$-th position of the vector is set to 1, otherwise it is set to 0. A total of $K_{p}$ meta-features are extracted using output profiles.  

\item \emph{\boldsymbol{$f_{5}$} \textbf{- Classifier's confidence:}} The perpendicular distance between the reference sample $\mathbf{x}_{j}$ and the decision boundary of the base classifier $c_{i}$ is calculated and encoded as $f_{5}$. 

\end{itemize}

A vector $v_{i,j} = \left\lbrace f_{1} \cup f_{2} \cup f_{3} \cup f_{4} \cup f_{5} \right\rbrace$ is obtained at the end of the process. If $c_{i}$ correctly classifies $\mathbf{x}_{j}$, the class attribute of $v_{i,j}$, $\alpha_{i,j} = 1$ (i.e., $v_{i,j}$ belongs to the meta-class ``competent''), otherwise $\alpha_{i,j} = 0$. $v_{i,j}$ is stored in the meta-features dataset $\mathcal{T}_{\lambda}^{*}$ that is used to train the meta-classifier $\lambda$.  

\subsubsection{Training}

The last step of the meta-training phase is the training of $\lambda$. The dataset $\mathcal{T}_{\lambda}^{*}$ is divided on the basis of 75\% for training and  25\% for validation. In this paper, we evaluate four classifier models for the meta-classifier: MLP Neural Network, Support Vector Machines with Gaussian Kernel (SVM), Random Forests and Naive Bayes. These classifiers were selected based on a recent study~\cite{delgado14a} that ranked the best classification models in a comparison considering a total of 179 classifiers and 121 datasets. All classifiers were implemented using the Matlab PRTOOLS toolbox~\cite{PRTools}. The parameters of each classifier were set as follows:

\begin{enumerate}

\item MLP Neural Network: The validation data was used to select the number of nodes in the hidden layer. We used a configuration with 10 neurons in the hidden layer since there were no improvement in results with more than 10 neurons. The training process for $\lambda$ was performed using the Levenberg-Marquadt algorithm. The training process was stopped if its performance on the validation set decreased or failed to improve for five consecutive epochs.

\item SVM: A radial basis SVM with a Gaussian Kernel was used. For each dataset, a grid search was performed in order to set the values of the regularization parameter $c$ and the Kernel spread parameter $\gamma$. 

\item Random Forest: A total of $200$ decision trees were used. The depth of each tree was fixed at $5$. 

\item Naive Bayes: A simple Naive Bayes classifier using a normal distribution to model numeric features. No parameters are required for this model.

\end{enumerate}

\subsection{Generalization}
\label{sec:generalization}

Given the query sample $\mathbf{x}_{j,test}$, the region of competence $\theta_{j}$ is computed using the samples from the dynamic selection dataset $D_{SEL}$. Following that, the output profiles $\tilde{\mathbf{x}}_{j,test}$ of the test sample, $\mathbf{x}_{j,test}$, are calculated. The set with $K_{p}$ similar output profiles $\phi_{j}$, of the query sample $\mathbf{x}_{j,test}$, is obtained through the Euclidean distance applied over the output profiles of the dynamic selection dataset, $\tilde{D}_{SEL}$.

Next, for each classifier $c_{i}$ belonging to the pool of classifiers $C$, the meta-features extraction process is called, returning the meta-features vector $v_{i,j}$. Then, $v_{i,j}$ is used as input to the meta-classifier $\lambda$. The support obtained by the meta-classifier for the ``competent'' meta-class, denoted by $\delta_{i,j}$, is computed as the level of competence of the base classifier $c_{i}$ for the classification of the test sample $\mathbf{x}_{j,test}$. 

Three combination approaches are considered:

\begin{itemize}

\item \textbf{META-DES.S}: In this approach, the base classifiers that achieve a level of competence $\delta_{i,j} > \Upsilon$ are considered competent, and are selected to compose the ensemble $C'$. In this paper, we set $\Upsilon = 0.5$ (i.e., the base classifier is selected if the support for the "`competent"' meta-class is higher than the support for the "`incompetent"' meta-class). The final decision is obtained using the majority vote rule~\cite{kittler}. Tie-breaking is handled by choosing the class with the highest a posteriori probability.

\item \textbf{META-DES.W}: Every classifier in the pool $C$ is used to predict the label of $\mathbf{x}_{j,test}$. The level of competence $\delta_{i,j}$ estimated by the meta-classifier $\lambda$ is used as the weight of each base classifier. The final decision is obtained using a weighted majority vote combination scheme~\cite{kuncheva}. Thus, the decisions obtained by the base classifiers with a higher level of competence $\delta_{i,j}$ have a greater influence on the final decision.

\item \textbf{META-DES.H}: In this  approach, first the base classifiers that achieve a level of competence $\delta_{i,j} > \Upsilon = 0.5$ are considered competent and are selected to compose the ensemble $C'$. Next, the level of competence $\delta_{i,j}$ estimated by the meta-classifier $\lambda$, for the classifiers in the ensemble $C'$, are used as its weights. Thus, the decisions obtained by the base classifiers with the highest level of competence $\delta_{i,j}$ have a greater influence in the final decision. A weighting majority voting scheme is used to predict the label $w_{l}$ of $\mathbf{x}_{j,test}$. 

\end{itemize}
  
\section{Experiments}
\label{sec:experiments}

\subsection{Datasets}
 
A total of 30 datasets are used in the comparative experiments, with sixteen taken from the UCI machine learning repository~\cite{Lichman2013}, four from the STATLOG project~\cite{King95statlog}, four from the Knowledge Extraction based on Evolutionary Learning (KEEL) repository~\cite{FdezFLDG11}, four from the Ludmila Kuncheva Collection of real medical data~\cite{lkc}, and two artificial datasets generated with the Matlab PRTOOLS toolbox~\cite{PRTools}. The key features of each dataset are shown in Table~\ref{table:datasets}.
 
 \begin{table}[htbp]
     \centering
     \caption{Key Features of the datasets used in the experiments}
      \label{table:datasets} 
      \resizebox{0.50\textwidth}{!}{
     \begin{tabular}{l c c c c}
     \hline
       \textbf{Database} & \textbf{No. of Instances} & \textbf{Dimensionality} & \textbf{No. of Classes} & \textbf{Source}\\
         \hline
 
         \textbf{Pima} & 768 & 8 & 2 & UCI \\
 	
         \textbf{Liver Disorders} & 345 & 6 & 2 & UCI  \\
 
         \textbf{Breast (WDBC)} & 568 & 30 & 2 & UCI \\
 	
         \textbf{Blood transfusion} & 748 & 4 &	2 & UCI  \\
 
 	 \textbf{Banana}  & 1000 & 2 &	2 & PRTOOLS  \\
 
         \textbf{Vehicle} & 846 & 18 & 4 & STATLOG \\
 
 	\textbf{Lithuanian}  & 1000 & 2 & 2 & PRTOOLS  \\
 
         \textbf{Sonar} & 208 &	60 & 2 & UCI \\
  
         \textbf{Ionosphere} & 315 &	34 & 2 & UCI  \\
 
         \textbf{Wine} & 178 & 13 & 3 & UCI  \\
  
         \textbf{Haberman's Survival} & 306 & 3 & 2 & UCI  \\
 
        \textbf{Cardiotocography (CTG)} & 2126 & 21 & 3 & UCI \\    
 
        \textbf{Vertebral Column} & 310 & 6 & 2 & UCI  \\          

        \textbf{Steel Plate Faults} & 1941 & 27 & 7 & UCI \\   
 
        \textbf{WDG V1} & 50000 & 21 & 3 & UCI  \\    
 
        \textbf{Ecoli} & 336 & 7 & 8 & UCI   \\    
 	   
        \textbf{Glass} & 214 & 9 & 6  & UCI  \\                       
 		
        \textbf{ILPD} & 214 & 9 & 6  & UCI  \\                       
 
         \textbf{Adult} & 48842 & 14 & 2 & UCI \\        
 
        \textbf{Weaning} & 302 & 17 & 2 & LKC \\

         \textbf{Laryngeal1} & 213 & 16 & 2 & LKC  \\        
 
        \textbf{Laryngeal3} & 353 & 16 & 3 & LKC \\    
 
        \textbf{Thyroid} &  215 & 5 & 3 & LKC  \\
      
         \textbf{German credit} & 1000 & 20 &2  & STATLOG  \\
 
         \textbf{Heart} & 270 & 13  & 2  & STATLOG  \\
         
         \textbf{Satimage} & 6435 & 19 & 7 & STATLOG   \\    
 
         \textbf{Phoneme} & 5404 & 6 & 2 & ELENA \\   
 
		\textbf{Monk2}  & 4322 & 6 & 2 & KEEL  \\
 		          
 		\textbf{Mammographic}  & 961 & 5 & 2 & KEEL  \\
		
     	\textbf{MAGIC Gamma Telescope}  & 19020 & 10 & 2 & KEEL  \\
 	    \hline  
     \end{tabular}
 		}
 \end{table}

\begin{table*}[ht!] 
\centering 
\caption{Comparison of different classifier types used as the meta-classifier $\lambda$ for the META-DES framework. The best results are in bold. Results that are significantly better are marked with a $\bullet$.} 
\label{table:diffClassifiers}  
\resizebox{1.0 \textwidth}{!}{  
\begin{tabular}{l |c| c c c | c | c c c|}  
\cline{2-9}

 & \multicolumn{4}{ |c| }{\textbf{Meta-Classifier $\lambda$}} & \multicolumn{4}{ |c| }{\textbf{META-DES}} \\ \hline
 
\multicolumn{1}{|l|}{ \textbf{Dataset} } & $\lambda$ \textbf{MLP NN} & $\lambda$ \textbf{SVM} & $\lambda$ \textbf{Forest} & $\lambda$ \textbf{Bayes} & \textbf{MLP NN} & \textbf{SVM} & \textbf{Forest } & \textbf{Bayes} \\ 
 \hline

\multicolumn{1}{|l|}{ \textbf{Pima} } & 78.53(1.24) & 79.46(1.67) & \textbf{80.27(2.08)} & 79.63(1.75) & \textbf{79.03(2.24)} & 77.58(1.67) & 78.39(2.08) & 77.76(1.75) \\

\multicolumn{1}{|l|}{ \textbf{Liver} }& 68.83 (5.57)  & 70.60(5.52) & 69.56(5.17) & \textbf{71.24(4.84)} & \textbf{70.08(3.49)} & 68.92(5.52) & 67.88(5.17) & 69.56(4.84) \\
 
\multicolumn{1}{|l|}{ \textbf{Breast} } & 95.43 (1.85)  & 97.19(0.61) & 97.19(0.61) & \textbf{97.66(0.50)} & \textbf{97.41(1.07)} & 96.94(0.61) & 96.94(0.61) & 96.94(0.61) \\
 
\multicolumn{1}{|l|}{ \textbf{Blood} } & 79.54(3.03) & 79.18(1.88) & \textbf{79.83(2.42)} & 79.66(1.52)  & \textbf{79.14(1.03)} & 77.84(1.88) & 78.49(2.42) & 78.31(1.52) \\
 
\multicolumn{1}{|l|}{ \textbf{Banana} } & 91.14(3.09)   & 95.17(1.75) & 90.97(3.89) & \textbf{95.67(2.37)} $\bullet$ & 91.78(2.68) & 93.92(1.75) & 89.72(3.89) & \textbf{94.42(2.37)} $\bullet$ \\ 
 
\multicolumn{1}{|l|}{ \textbf{Vehicle} } & 82.38(2.34)  & 82.50(1.92) & 82.44(1.63) & \textbf{82.76(2.01)}  & 82.75(1.70) & 83.29(1.92) & 83.24(1.63) & \textbf{83.55(2.01)} \\

\multicolumn{1}{|l|}{ \textbf{Lithuanian} } & 93.42(3.41)  & 94.91(1.25) & \textbf{97.89(0.81)} $\bullet$ & 93.72(3.09) & 93.18(1.32) & 94.30(1.25) & \textbf{97.28(0.81)} $\bullet$ & 93.12(3.09) \\
 
\multicolumn{1}{|l|}{ \textbf{Sonar} } & 86.15(2.43)  & 85.88(4.08) & 84.60(4.61) & \textbf{86.95(5.67)} $\bullet$ & 80.55(5.39) & 80.77(4.08) & 79.49(4.61) & \textbf{81.84(5.67)} \\

\multicolumn{1}{|l|}{ \textbf{Ionosphere} } & \textbf{89.18(2.31)} & 87.35(2.42) & 87.09(2.48) & 87.35(2.21) & \textbf{89.94(1.96)} & 89.06(2.42) & 88.80(2.48) & 89.06(2.21) \\

\multicolumn{1}{|l|}{ \textbf{Wine} } & \textbf{98.90(1.61)} & \textbf{98.90(1.61)} & \textbf{98.90(1.61)} & 97.25(1.48) & 99.25(1.11) & \textbf{99.27(1.61)} & 99.02(1.61) & 98.53(1.48) \\

\multicolumn{1}{|l|}{ \textbf{Haberman} } & \textbf{76.31(2.35)} & 74.81(2.50) & 75.69(2.19) & 75.25(2.06) & \textbf{76.71(1.86)} & 75.69(2.50) & 76.56(2.19) & 76.13(2.06) \\

\multicolumn{1}{|l|}{ \textbf{CTG} } & 82.00(5.22)  & 88.81(1.03) & 88.60(1.04) & \textbf{90.21(1.14)} $\bullet$ & 84.62(1.08) & 85.64(1.03) & 85.43(1.04) & \textbf{86.04(1.14)} $\bullet$ \\ 

\multicolumn{1}{|l|}{ \textbf{Vertebral} } & 86.89(2.46) & 87.70(2.87) & \textbf{87.85(3.54)} & 86.56(2.35) & 86.89(2.46) & 86.76(2.87) & \textbf{86.90(3.54)} & 85.62(2.35) \\

\multicolumn{1}{|l|}{ \textbf{Faults} } & 70.21(4.26) & 74.41(1.17) & 74.41(1.17) & \textbf{74.68(1.19)} $\bullet$ & 67.21(1.20) & 68.45(1.17) & 68.45(1.17) & \textbf{68.72(1.19)} $\bullet$ \\

\multicolumn{1}{|l|}{ \textbf{WDVG1} } & 83.26(1.36) & 85.26(0.63) & 85.23(0.50) & \textbf{85.84(0.60)} $\bullet$ & 84.56(0.36) & 84.67(0.63) & 84.64(0.50) & \textbf{84.84(0.36)} $\bullet$ \\

\multicolumn{1}{|l|}{ \textbf{Ecoli} } & 77.09(4.84) & \textbf{78.01(3.89)} & 76.74(3.58) & 77.01(3.76) & 77.25(3.52) & \textbf{80.92(3.89)} $\bullet$ & 80.66(3.58) & 80.92(3.76) \\
 
\multicolumn{1}{|l|}{ \textbf{GLASS} } & \textbf{69.18(1.49)} $\bullet$ & 63.31(4.40) & 64.84(4.44) & 64.89(3.65) & \textbf{66.87(2.99)} $\bullet$ & 65.62(4.40) & 64.16(4.44) & 65.21(3.65) \\

\multicolumn{1}{|l|}{ \textbf{ILPD} } & 69.80(4.96) & 70.48(2.17) & 69.95(2.32) & \textbf{71.09(2.33)} $\bullet$ & 69.40(1.64) & 69.56(2.17) & 69.03(2.32) & \textbf{70.17(2.33)} \\

\multicolumn{1}{|l|}{ \textbf{Adult} } & 87.00(6.29) & \textbf{88.75(1.76)} & 88.68(1.29) & 88.62(1.84) & 87.15(2.43) & \textbf{87.35(1.76)} & 87.29(1.29) & 87.22(1.84) \\
 
\multicolumn{1}{|l|}{ \textbf{Weaning} } & 79.55(4.44)  & 79.75(2.85) & 79.75(2.85) & \textbf{80.33(3.71)} & 79.67(3.78) & 79.10(2.85) & 79.10(2.85) & \textbf{79.69(3.71)} \\

\multicolumn{1}{|l|}{ \textbf{Laryngeal1} } & 77.81(3.51) & 80.08(3.67) &  \textbf{81.29(3.79)} $\bullet$ & 79.94(5.00) & 79.67(3.78) & 81.97(3.67) & \textbf{82.18(3.78)} $\bullet$ & 81.97(5.00) \\
 
\multicolumn{1}{|l|}{ \textbf{Laryngeal3} } & 72.42(3.57) & 72.63(0.87) & 72.76(0.81) & \textbf{73.82(0.67)}  & 72.65(2.17) & 73.17(2.32) & 74.04(2.23) & \textbf{74.42(1.26)} $\bullet$ \\

\multicolumn{1}{|l|}{ \textbf{Thyroid} } & 96.16(5.96)  & 97.27(2.32) & 97.15(2.23) & \textbf{97.52(1.26)} & 96.78(0.87) & 97.18(0.87) & 97.31(0.81) & \textbf{97.38(0.67)} \\

\multicolumn{1}{|l|}{ \textbf{German} } & 75.00(4.18)  & 76.18(2.82) & \textbf{77.11(1.58)} $\bullet$ & 75.38(1.30) & 75.55(1.31) & 75.34(2.82) & \textbf{76.27(2.58)} & 74.54(1.30) \\

\multicolumn{1}{|l|}{ \textbf{Heart} } & 84.38(4.63)  & 83.67(2.76) & 82.85(3.60) & \textbf{86.99(2.30)} $\bullet$ & 84.80(3.36) & 84.97(2.76) & 84.15(3.60) & \textbf{85.30(2.30)} \\
 
\multicolumn{1}{|l|}{ \textbf{Segmentation} } & 96.89(0.74)  & 96.78(0.60) & 96.95(0.75) & \textbf{96.99(0.60)}  & 96.21(0.87) & 96.21(0.60) & 96.38(0.75) & \textbf{96.42(0.76)} \\

\multicolumn{1}{|l|}{ \textbf{Phoneme} } & 80.99(3.88) & 86.80(3.19) & 86.80(3.19) & \textbf{90.13(0.72)} $\bullet$ & 80.35(2.58) & 78.44(3.19) & 78.44(3.19) & \textbf{81.77(0.72)} \\
 
\multicolumn{1}{|l|}{ \textbf{Monk2} } & 83.89(2.59)  & 86.40(2.82) & 85.68(2.45) & \textbf{88.67(3.32)} $\bullet$ & 83.24(2.19) & 81.08(2.82) & 80.36(2.45) & \textbf{83.34(3.32)} \\

\multicolumn{1}{|l|}{ \textbf{Mammographic} } & 78.00(5.93)  & \textbf{87.30(1.82)} $\bullet$ & 87.30(1.53) & 86.34(2.54) & 84.82(1.55) & \textbf{85.37(1.82)} & 85.37(1.53) & 84.41(2.54) \\

\multicolumn{1}{|l|}{ \textbf{Magic Gamma Telescope} } & 75.40(2.25) & 72.30(3.33) & 74.57(3.56) & \textbf{78.65(2.52)} & 84.35(3.27) & 81.35(4.21) & 84.35(3.27)  & \textbf{85.33(2.29)}  \\

\hline

\multicolumn{1}{|l|}{ \textbf{Wilcoxon Signed Test} } & n/a & \textasciitilde $\: (\rho = 0.110)$ & $+$ $(\rho = 0.004)$ & $+$ $(\rho = 0.007)$ & n/a & \textasciitilde $\: (\rho = 0.70)$ & \textasciitilde $\: (\rho = 0.500)$ & \textasciitilde $\: (\rho = 0.30)$  \\

\hline 
\end{tabular} } 
\end{table*}

\subsection{Experimental Protocol}

For the sake of simplicity, the same experimental protocol used in previous publications~\cite{CruzPR,icpr2014} was used. The experiments were carried out using 20 replications. For each replication, the datasets were randomly divided on the basis of 50\% for training, 25\% for the dynamic selection dataset (D$_{SEL}$), and 25\% for the test set ($\mathcal{G}$). The divisions were performed while maintaining the prior probabilities of each class. For the proposed META-DES, 50\% of the training data was used in the meta-training process $\mathcal{T}_{\lambda}$ and 50\% for the generation of the pool of classifiers ($\mathcal{T}$). 
 
For the two-class classification problems, the pool of classifiers was composed of 100 Perceptrons generated using the bagging technique~\cite{bagging}. For the multi-class problems, the pool of classifiers was composed of 100 multi-class Perceptrons. The use of Perceptron as base classifier is based on the observations that the use of weak classifiers can show more differences between the DES schemes~\cite{knora}, thus making it a better option for comparing different DES techniques. Furthermore, as reported by Leo Breiman, the bagging technique achieves better results when weak and unstable base classifiers are used~\cite{bagging}.

The values of the hyper-parameters $K$, $K_{p}$ and $h_{c}$ were set at 7, 5 and 70\%, respectively. They were selected empirically based on previous publications~\cite{ijcnn2011,icpr2014,CruzPR}.

\subsection{Comparison of different classification models as the Meta-Classifier}
\label{sec:diffClassifiers}

In this experiment, we analyze the impact of the classifier model used for the meta-problem (i.e., for the selection of competent classifiers). The objective of this experiment is to verify whether we can improve the classification performance of the META-DES system, previously defined using an MLP neural network as the meta-classifier. The following classifier models are considered: Multi-Layer Perceptron (MLP) Neural Networks as in~\cite{CruzPR}, Support Vector Machines with Gaussian Kernel (SVM), Random Forests and Naive Bayes.

Table~\ref{table:diffClassifiers} shows a comparison of the performance of the meta-classifier $\lambda$ and the recognition accuracy obtained by the META-DES system using each classification model. The best results are highlighted in bold. For each dataset, we compared the results obtained by the meta-classifier $\lambda$ and by the META-DES framework using the MLP network~\cite{CruzPR}, against the best result obtained by any of the other classifier models (SVM, Random Forest and Naive Bayes). The comparison was performed using the Kruskal-Wallis non-parametric statistical test, with a 95\% confidence interval. Results that are significantly better are marked with a $\bullet$.

\begin{figure}[!ht]
  
   \begin{center}  	 
       	  \epsfig{file=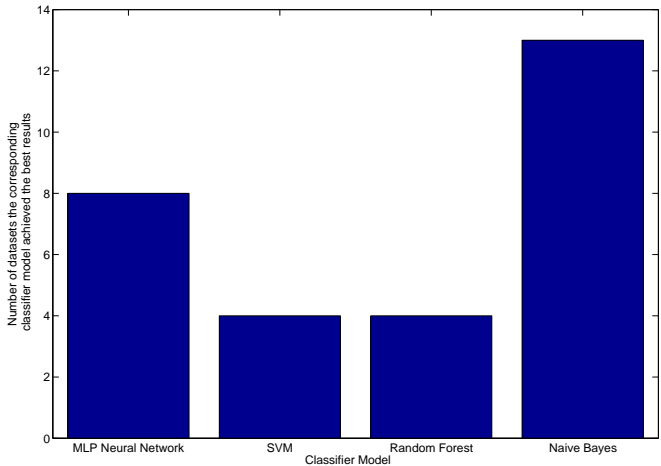, clip=,  width=0.50\textwidth}
   \end{center}
\caption{Bar plot showing the number of datasets that each classification model used a the meta-classifier $\lambda$ presented the highest recognition accuracy.}
\label{fig:barClassifierModel}
\end{figure}

We can observe that when the meta-classifier achieves a recognition performance that is statistically superior for a single dataset, such as, Banana, Faults and WDGV1, for instance, the META-DES is also likely to achieve superior accuracy for the same classification problem. Figure~\ref{fig:barClassifierModel} shows the number of datasets that each classifier model achieved the highest accuracy. The Naive Bayes classifier is ranked first, achieving the best results for 14 datasets, followed by the MLP Neural Network with 8. SVM and Random Forests achieved the best results for 4 datasets each. The strong performance of the Naive Bayes may be explained by the fact that the majority of the meta-features are binary, and this classifier model handles well binary input features different than MLP Networks. In addition, it might indicate that the proposed sets of meta-features are possibly independent~\cite{Cruz2014ANNPR}. This is an interesting finding since the Naive Bayes model is much faster both in the training and testing stages when compared to an MLP Neural Network or an SVM classifier. 

Furthermore, in order to verify whether the difference in classification results obtained over the 30 datasets is statistically significant, we performed a Wilcoxon non-parametric signed rank test with 95\% confidence for a pairwise comparison between the results obtained using an MLP Neural Network against the best result obtained using a different classifier for the meta-classifier. The Wilcoxon signed rank test was used since it was suggested in~\cite{Demsar:2006} as a robust method for comparing the classification results of two algorithms over several datasets. The results of the Wilcoxon statistical test are shown in the last row of Table~\ref{table:diffClassifiers}. Techniques that achieve performance equivalent to the MLP network are marked with "\textasciitilde"; those that achieve statistically superior performance are marked with a "+", and those with inferior performance are marked with a "-". When comparing the performance of the four meta-classifiers, the results achieved using Random Forests and Naive Bayes as the meta-classifier $\lambda$ are significantly superior. 

Hence, we can conclude that significant gains in classification accuracy can be achieved by choosing a more suitable classifier model for the meta-classifier $\lambda$. Although the choice of the best meta-classifier may vary according to the classification problem (Table~\ref{table:diffClassifiers}), the results of the META-DES using Naive Bayes as the meta-classifier achieves results that are statistically superior when compared to the MLP neural network over the 30 datasets studied in this work. 

\subsection{Comparison Between Combination Approaches: Dynamic Selection, Dynamic Weighting and Hybrid}
\label{sec:fusion}

In this section, we compare the three combination approaches presented in Section~\ref{sec:generalization}: Dynamic Selection, Dynamic weighting, and the Hybrid approach. For the sake of simplicity, we present only the results obtained using the Naive Bayes as the meta-classifier $\lambda$ since it achieved the highest classification accuracy in the previous experiments (Table~\ref{table:diffClassifiers}). 

The results achieved using the Naive Bayes as meta-classifier for the three combination approaches are shown in Table~\ref{table:classificationApproaches}. 
In order to select the best combination approach, we compare the average ranks of each approach computed using the Friedman test, which is a non-parametric equivalent of the repeated measures ANOVA used to compare several algorithms over multiple datasets~\cite{Friedman1937Use,Demsar:2006}. The Friedman test ranks each algorithm, with the best performing one getting rank 1, the second best rank 2, and so forth for each dataset separately. The average rank is then computed, considering all datasets. Thus, the best algorithm is the one with the lowest average rank. The approaches that use the proposed weighting scheme (Dynamic weighting and Hybrid) outperformed the Dynamic selection approach in accuracy. This can be explained by the fact the outputs given by the Naive Bayes classifier can be directly interpreted as the likelihood that the base classifier belongs to the "`competent"' meta-class. Thus, the supports provided by the meta-classifier can directly be used as the weights of each classifier for a weighted majority voting scheme. This is different from other classification models, such as Random Forests where their class supports cannot be directly interpreted as such. Hence, the meta-classifier can also be used for the fusion (integration) of the classifiers in the ensemble, rather than only for ensemble selection. Since the Hybrid combination approach presents the highest recognition accuracy when the 30 datasets are considered (lowest average rank) this combination approach is selected for the comparison against other state-of-the-art DES techniques.

\begin{table}[ht!] 
\centering 
\caption{Comparison between the three classification approaches: Selection, Weighting and Hybrid for the META-DES framework. The results using a Naive Bayes as the meta-classifier $\lambda$ are presented. The best results are in bold. The average rank is shown in the last row of the table.} 
\label{table:classificationApproaches}  
\resizebox{0.50	\textwidth}{!}{  
\begin{tabular}{l c c c}  
\hline  
Dataset & META-DES.S & META-DES.W & META-DES.H \\ 
\hline   
\textbf{Pima} & 77.76(1.75) & 77.64(1.68) & \textbf{77.93(1.86)}  \\ 

\textbf{Liver} & 69.56(4.84) & 69.69(4.68) & \textbf{69.95(3.49)} \\ 

\textbf{Breast} & \textbf{97.41(0.50)} & 97.25(0.47) & 97.25(0.47)  \\ 

\textbf{Blood} & 78.31(1.52) & \textbf{78.67(1.77)} & 78.25(1.37)  \\ 

\textbf{Banana} & 94.42(2.37) & \textbf{95.13(1.88)} & 94.51(2.36)  \\ 

\textbf{Vehicle} & \textbf{83.55(2.01)} & 83.50(1.87) & 83.55(2.10)  \\
 
\textbf{Lithuanian} & 93.12(3.09) & 93.19(3.14) & \textbf{93.26(3.22)}  \\
 
\textbf{Sonar} & 81.84(5.67) & 79.92(5.16) & \textbf{82.06(2.09)}  \\ 

\textbf{Ionosphere} & \textbf{89.06(2.21)} & 89.06(2.55) & 89.06(2.21)  \\
 
\textbf{Wine} & \textbf{98.53(1.48)} & \textbf{98.53(1.08)} & \textbf{98.53(1.08)}  \\
 
\textbf{Haberman} & 76.13(2.06) & \textbf{76.42(2.38)} & 76.13(1.56)  \\ 

\textbf{CTG} & \textbf{86.04(1.14)} & 85.99(1.05) & \textbf{86.08(1.24)}  \\ 

\textbf{Vertebral} & 85.62(2.35) & \textbf{85.76(2.55)} & 84.90(2.95)  \\
 
\textbf{Faults} & 68.72(1.19) & 68.63(1.24) & \textbf{68.95(1.04)}  \\ 

\textbf{WDVG1} & \textbf{84.84(0.36)} & 84.83(0.63) & 84.77(0.65)  \\ 

\textbf{Ecoli} & \textbf{80.92(3.76)} & 80.66(3.58) & 80.66(3.48)  \\ 

\textbf{GLASS} & 65.21(3.65) & \textbf{66.04(3.67)} & 65.21(3.53)  \\ 

\textbf{ILPD} & 70.17(2.33) & \textbf{70.48(2.28)} & 69.64(2.47)  \\ 

\textbf{Adult} & 87.22(1.84) & 87.29(2.20) & \textbf{87.29(1.80)}  \\ 

\textbf{Weaning} & 79.69(3.71) & 79.83(2.94) & \textbf{79.98(3.55)}  \\ 

\textbf{Laryngeal1} & 87.00(5.00) & 86.79(4.72) & \textbf{87.21(5.35)}  \\

\textbf{Laryngeal3} & 73.42(1.26) & \textbf{73.79(1.38)} & 73.54(1.66)  \\ 

\textbf{Thyroid} & 97.38(0.67) & \textbf{97.44(0.71)} & 97.38(0.67)  \\ 

\textbf{German} & 74.54(1.30) & \textbf{75.03(2.04)} & 74.36(1.28)  \\
 
\textbf{Heart} & 85.30(2.30) & \textbf{85.46(2.70)} & 85.46(2.70)  \\ 

\textbf{Segmentation} & 96.42(0.76) & 96.34(0.74) & \textbf{96.46(0.79)}  \\ 

\textbf{Phoneme} & 81.77(0.72) & 81.47(0.77) & \textbf{81.82(0.69)}  \\ 

\textbf{Monk2} & 83.34(3.32) & 82.83(3.82) & \textbf{83.45(3.46)}  \\ 

\textbf{Mammographic} & 84.41(2.54) & \textbf{84.62(2.46)} & 84.30(2.27)  \\ 

\textbf{Magic Gamma Telescope} & 85.33(2.29) & 84.62(2.46) & \textbf{85.65(2.27)}  \\ 

\hline

\textbf{Friedman Average Rank ($\downarrow$)} & 2.15 & 1.98 & \textbf{1.86}  \\ 

\hline 
\end{tabular} } 
\end{table}

\subsection{Comparison with the state-of-the-art DES techniques}

In this section, we compare the recognition rates obtained by the proposed META-DES.H against eight state-of-the-art dynamic selection techniques in the DES literature: the KNORA-ELIMINATE~\cite{knora}, KNORA-UNION~\cite{knora}, DES-FA~\cite{ijcnn2011}, Local Classifier Accuracy (LCA)~\cite{lca}, Overall Local Accuracy (OLA)~\cite{lca}, Modified Local Accuracy (MLA)~\cite{Smits_2002}, Multiple Classifier Behaviour (MCB)~\cite{mcb} and K-Nearests Output Profiles (KNOP)~\cite{paulo2}. 

For all techniques, we use the same pool of classifiers defined in the previous section (Section~\ref{sec:diffClassifiers}) in order to have a fair comparison. The size of the region of competence (neighborhood size), $K$ is set to $7$ since it achieved the best result in previous experiments~\cite{Alceu2014,ijcnn2011}. The comparative results are shown in Table~\ref{table:ResultsBayes}. Due to size constraints, we only show the results using Naive Bayes as the meta-classifier since it achieved the highest recognition accuracy in the previous experiment. For each dataset, a Kruskal-Wallis statistical test with 95\% confidence was conducted to know if the classification improvement is statistically significant. Results that are statistically better are marked with a $\bullet$. The results of the proposed technique obtained the highest accuracy in 20 out of 30 datasets. In addition, the accuracy of the proposed system was statistically superior in 15 out of 30 datasets. The original META-DES framework~\cite{CruzPR}, without the improvements proposed in this paper, achieved results that are statistically superior in 10 out of the 30 datasets when compared with the state-of-the-art DES techniques. 

Furthermore, we also consider the Wilcoxon test with 95\% confidence, for a pairwise comparison between the classification performances of the proposed system against the performance of the state-of-the-art DES techniques over multiple datasets. The results of the Wilcoxon test are shown in the last row of the table. The performance of the proposed META-DES.H system is statistically better when all 30 datasets are considered. Hence, the experimental results demonstrate that the changes proposed in this paper lead to a significant gains in performance when compared to other DES algorithms.

 \begin{table*}[ht]
     \centering
     \caption{Mean and standard deviation results of the accuracy obtained for the proposed META-DES using a Naive Bayes classifier for the meta-classifier $\lambda$ and the hybrid combination approach. The best results are in bold. Results that are significantly better are marked with a $\bullet$.}
      \label{table:ResultsBayes} 
      \resizebox{1.0\textwidth}{!}{
      \begin{tabular}{|l  c  c  c  c  c  c  c  c  c|}
     \hline
        \textbf{Database} & \textbf{META-DES.H} & \textbf{KNORA-E}~\cite{knora} & \textbf{KNORA-U}~\cite{knora} & \textbf{DES-FA}~\cite{ijcnn2011} & \textbf{LCA}~\cite{lca} & \textbf{OLA}~\cite{lca} & \textbf{MLA}~\cite{Smits_2002} & \textbf{MCB}~\cite{mcb} &  \textbf{KNOP}~\cite{paulo2} \\
         \hline
 
          \textbf{Pima} & \textbf{77.93(1.86)}  &  73.79(1.86) &  76.60(2.18) & 73.95(1.61) & 73.95(2.98)  & 73.95(2.56) & 77.08(4.56) & 76.56(3.71) & 73.42(2.11) \\

         \textbf{Liver Disorders} & \textbf{69.95(3.49)} $\bullet$ &  56.65(3.28) &  56.97(3.76) & 61.62(3.81) & 58.13(4.01)  & 58.13(3.27) & 58.00(4.25)  & 58.00(4.25) & 65.23(2.29) \\

         \textbf{Breast (WDBC)} & 97.25(0.47) & 97.59(1.10)  &  97.18(1.02) & \textbf{97.88(0.78)} & 97.88(1.58)  & 97.88(1.58)  & 95.77(2.38)  & 97.18(1.38) & 95.42(0.89) \\

         \textbf{Blood Transfusion} & \textbf{78.25(1.37)} $\bullet$ &  77.65(3.62) &  77.12(3.36) & 73.40(1.16) & 75.00(2.87) & 75.00(2.36) & 76.06(2.68)  & 73.40(4.19) & 77.54(2.03) \\

         \textbf{Banana} &  94.51(2.36)  &  93.08(1.67) & 92.28(2.87) & \textbf{95.21(3.18)} & 95.21(2.15)  & 95.21(2.15)  & 80.31(7.20) &  88.29(3.38) & 90.73(3.45) \\

         \textbf{Vehicle} & 83.55(2.10)  &  83.01(1.54) &  82.54(1.70)  & 82.54(4.05) & 80.33(1.84)  & 81.50(3.24) & 74.05(6.65) & \textbf{84.90(2.01)} & 80.09(1.47) \\

         \textbf{Lithuanian Classes} & 93.26(3.22) &  93.33(2.50) &  95.33(2.64) & 98.00(2.46) & 85.71(2.20) & \textbf{98.66(3.85)} & 88.33(3.89)  & 86.00(3.33) & 89.33(2.29) \\

         \textbf{Sonar} & \textbf{82.06(2.09)} $\bullet$  &  74.95(2.79) &  76.69(1.94) & 78.52(3.86) & 76.51(2.06) & 74.52(1.54) & 76.91(3.20)  & 76.56(2.58) & 75.72(2.82) \\
         
         \textbf{Ionosphere} & 89.06(2.21) &  \textbf{89.77(3.07)} &  87.50(1.67) & 88.63(2.12) & 88.00(1.98)  & 88.63(1.98) & 81.81(2.52)  & 87.50(2.15) & 85.71(5.52) \\

         \textbf{Wine} & \textbf{98.53(1.08)} $\bullet$ &  97.77(1.53) &  97.77(1.62) &  95.55(1.77)  & 85.71(2.25)  & 88.88(3.02) & 88.88(3.02)  & 97.77(1.62) & 95.50(4.14)\\

         \textbf{Haberman} & \textbf{76.13(1.56)} $\bullet$ &  71.23(4.16)  & 73.68(2.27)  & 72.36(2.41)  &  70.16(3.56) &  69.73(4.17) &  73.68(3.61)  &  67.10(7.65) & 75.00(3.40) \\        

       \textbf{Cardiotocography (CTG)} & 86.08(1.24)  & \textbf{86.27(1.57)} &  85.71(2.20) & \textbf{86.27(1.57)} &  86.65(2.35) &  86.65(2.35)  & 86.27(1.78) & 85.71(2.21) & 86.02(3.04) \\		

        \textbf{Vertebral Column} & 84.90(2.95) &  85.89(2.27) & \textbf{87.17(2.24)} &  82.05(3.20) &  85.00(3.25) &  85.89(3.74) &  77.94(5.80) & 84.61(3.95) & 86.98(3.21)\\		

        \textbf{Steel Plate Faults} & \textbf{68.95(1.04)} & 67.35(2.01) & 67.96(1.98)  & 68.17(1.59)  &  66.00(1.69) &  66.52(1.65)  & 67.76(1.54)  & 68.17(1.59) & 68.57(1.85)  \\	

        \textbf{WDG V1} & \textbf{84.77(0.65)} $\bullet$ &  84.01(1.10) &  84.01(1.10) & 84.01(1.10) & 80.50(0.56)  & 80.50(0.56) & 79.95(0.85)  & 78.75(1.35) & 84.21(0.45) \\	

        \textbf{Ecoli} &  \textbf{80.66(3.48)} &  76.47(2.76)  & 75.29(3.41) & 75.29(3.41) &  75.29(3.41)  &  75.29(3.41) &  76.47(3.06) & 76.47(3.06) & 80.00(4.25) \\	

        \textbf{Glass} & 65.21(3.53)  &  57.65(5.85) &  61.00(2.88) & 55.32(4.98) & 59.45(2.65) & 57.60(3.65) & 57.60(3.65) & \textbf{67.92(3.24)} & 62.45(3.65)\\	          

        \textbf{ILPD} & \textbf{69.64(2.47) } &  67.12(2.35) &  69.17(1.58) &  67.12(2.35) & 69.86(2.20) & 69.86(2.20) & 69.86(2.20) &  68.49(3.27) & 68.49(3.27)  \\	     

         \textbf{Adult} &  \textbf{87.29(1.80)} $\bullet$ &  80.34(1.57) &  79.76(2.26) & 80.34(1.57) & 83.58(2.32) &  82.08(2.42) & 80.34(1.32)  & 78.61(3.32) & 79.76(2.26) \\	 

        \textbf{Weaning} &  \textbf{79.98(3.55)}   & 78.94(1.25) & 81.57(3.65) & 82.89(3.52) & 77.63(2.35) & 77.63(2.35) & 80.26(1.52) & 81.57(2.86) & 82.57(3.33) \\	

         \textbf{Laryngeal1} & \textbf{87.21(5.35)} $\bullet$ &  77.35(4.45) & 77.35(4.45) &  77.35(4.45) &  77.35(4.45) & 77.35(4.45) &  75.47(5.55) & 77.35(4.45) & 77.35(4.45) \\	

        \textbf{Laryngeal3} &  \textbf{73.54(1.66)} & 70.78(3.68) & 72.03(1.89) & 72.03(1.89) &  72.90(2.30) & 71.91(1.01) & 61.79(7.80) & 71.91(1.01) & 73.03(1.89)  \\

        \textbf{Thyroid} & \textbf{97.38(0.67)} $\bullet$ & 95.95(1.25) &  95.95(1.25) &  95.37(2.02) & 95.95(1.25) & 95.95(1.25) &  94.79(2.30) &  95.95(1.25) & 95.95(1.25) \\	

         \textbf{German credit} & \textbf{74.54(0.30)} $\bullet$ & 72.80(1.95) & 72.40(1.80)  & 74.00(3.30) & 73.33(2.85) & 71.20(2.52) & 71.20(2.52) & 73.60(3.30)  & 73.60(3.30) \\	

         \textbf{Heart} & 85.46(2.70)  &  83.82(4.05) & 83.82(4.05) & 83.82(4.05) &  85.29(3.69) & 85.29(3.69) & \textbf{86.76(5.50)} & 83.82(4.05) & 83.82(4.05) \\	

         \textbf{Satimage} & \textbf{96.72(0.76)} $\bullet$ &  95.35(1.23) & 95.86(1.07) & 93.00(2.90) & 95.00(1.40) & 94.14(1.07) & 93.28(2.10) & 95.86(1.07) & 95.86(1.07) \\	

         \textbf{Phoneme} & \textbf{81.82(0.69)}  $\bullet$ &  79.06(2.50) &  78.92(3.33) &  79.06(2.50) & 78.84(2.53) &  78.84(2.53) &  64.94(7.75) & 73.37(5.55) & 78.92(3.33) \\	

 		\textbf{Monk2} & \textbf{83.45(3.46)}   $\bullet$ & 80.55(3.32) & 77.77(4.25) & 75.92(4.25)  & 74.07(6.60) & 74.07(6.60) & 75.92(5.65) & 74.07(6.60) & 80.55(3.32) \\	

 		\textbf{Mammographic} & \textbf{84.30(2.27)} $\bullet$ & 82.21(2.27) & 82.21(2.27) & 80.28(3.02) & 82.21(2.27 & 82.21(2.27) & 75.55(5.50) & 81.25(2.07) & 82.21(2.27) \\	

     	\textbf{MAGIC Gamma Telescope} &  \textbf{85.65(2.27)}  $\bullet$ &  80.03(3.25) & 79.99(3.55) & 81.73(3.27) & 81.53(3.35) & 81.16(3.00) & 73.13(6.35)  & 75.91(5.35) & 80.03(3.25) \\
     	
     	\hline
     	\textbf{Wilcoxon Signed test} &  n/a  & $-$ $(\rho = .0001)$ & $-$ $(\rho = .0007)$ & $-$ $(\rho = .0016)$ & $-$ $(\rho = .0001)$ & $-$ $(\rho = .0001)$ & $-$ $(\rho = .0001)$ & $-$ $(\rho = .0003)$ & $-$ $(\rho = .005)$ \\	

     \hline
     \end{tabular}
     }

 \end{table*}

\section{Conclusion}

In this paper, we proposed two modifications to the novel META-DES framework. First, we compared different classifier models, such as the MLP Neural Network, Support Vector Machines with Gaussian Kernel (SVM), Random Forests and Naive Bayes for the meta-classifier. Next, we evaluated three combination approaches to the framework: Dynamic selection, Dynamic weighting and Hybrid. In the Dynamic selection approach, only the classifiers that attain a certain level of competence are used to classify a given query sample. In the dynamic weighting approach, all base classifiers in the pool are considered to give the final decision, with the meta-classifier estimating the weight of each base classifier. In the hybrid approach, only the classifiers that attain a certain level of competence are initially selected, after which their decisions are aggregated in a weighted majority voting scheme. Thus, the base classifiers attaining higher levels of competence have a greater impact on the final decision.

Experiments were conducted using 30 classification datasets derived from five different data repositories (UCI, KEEL, STATLOG, LKC and ELENA). 
First, we observed a significant improvement in accuracy using different classifier models for the meta-problem. The performance of the META-DES trained using a Naive Bayes for the meta-classifier achieves results that are statistically better compared to those achieved using an MLP Neural Network, according to the Wilcoxon Signed Rank test with 95\% confidence. This finding confirms the initial hypothesis that the overall performance of the system improves when the recognition accuracy of the meta-classifier improves. As the META-DES framework considers the dynamic selection problem as a meta-classification problem, we can improve the recognition accuracy by focusing only on improving the classification performance in the meta-problem. This finding is especially useful for ill-defined problems since there is not enough data to properly train the base classifiers. Techniques such as stacked generalization for the generation of more meta-feature vectors in the data generation process as well as the use of feature selection techniques to achieve a more representative set of meta-features can be considered to improve the recognition performance at the meta-classification level.

In addition, we demonstrate that the framework can also be used to compute the weights of the base classifiers. We found that the Naive Bayes classifier achieved the best result when the dynamic weighting (META-DES.W) or hybrid (META-DES.H) approach is used. This can be explained by the fact that the supports given by this classifier can be seen as the likelihood that the base classifier belongs to the "`competent"' meta-class. Thus, the classifiers that are more likely to be "`competent"' have greater influence on the classification of any given test sample. When compared to eight state-of-the-art techniques found in the dynamic ensemble selection literature, the proposed META-DES.H using a Naive Bayes classifier for the meta-classifier presented classification accuracy that is statistically better in 15 out of the 30 classification datasets. The original META-DES framework~\cite{CruzPR} achieved results that are statistically better in 10 out of the 30 datasets when compared with the state-of-the-art DES techniques. Hence, the changes to the META-DES framework proposed in this paper lead to a significant gain in performance when compared against other DES algorithms. 

\section*{Acknowledgement}

This work was supported by the Natural Sciences and Engineering Research Council of Canada (NSERC), the \'{E}cole de technologie sup\'{e}rieure (\'{E}TS Montr\'{e}al), CNPq (Conselho Nacional de Desenvolvimento Cient\'{i}fico e Tecnol\'{o}gico) and FACEPE (Funda\c{c}\~{a}o de Amparo \`{a} Ci\^{e}ncia e Tecnologia de Pernambuco).
 
\bibliographystyle{IEEEtran}
\bibliography{report}

\end{document}